\pdfoutput=1
\documentclass[letterpaper, 10 pt, conference]{ieeeconf}  

\IEEEoverridecommandlockouts                              

\usepackage{cite}
\usepackage{amsmath,amssymb,amsfonts}
\usepackage{algorithmic}
\usepackage{graphicx}
\usepackage{textcomp}
\usepackage{xcolor}
\usepackage{multirow}
\usepackage{threeparttable}
\usepackage{array}
\usepackage{booktabs}
\usepackage{amsbsy}
\usepackage{ulem}
\usepackage{subfig}
\usepackage{float}
\usepackage{url}

\title{\LARGE \bf
Direct Sparse Odometry with Continuous 3D Gaussian Maps for Indoor Environments
}

\author{Jie~Deng$^{1}$, Fengtian~Lang$^{1}$ , Zikang~Yuan$^{2*}$and Xin~Yang$^{1*}$
	\thanks{$^{1}$Jie~Deng, Fengtian~Lang and Xin~Yang$^{*}$ are with the Electronic Information and Communications, Huazhong University of Science and Technology, Wuhan, 430074, China. (E-mail: {\tt\small d202481242@hust.edu.cn; m202372913@hust.edu.cn; xinyang2014@hust.edu.cn})}%
	\thanks{$^{2}$Zikang~Yuan$^{*}$ is with AI Chip Center for Emerging Smaert Systems, Hong Kong University of Science and Technology, Hong Kong. (* represents the corresponding author. E-mail: {\tt\small zikangyuan@ust.hk})}%
}

\begin{document}

\maketitle

\thispagestyle{empty}
\pagestyle{empty}

\begin{abstract}
	Accurate localization is essential for robotics and augmented reality applications such as autonomous navigation. Vision-based methods combining prior maps aim to integrate LiDAR-level accuracy with camera cost efficiency for robust pose estimation. Existing approaches, however, often depend on unreliable interpolation procedures when associating
	discrete point cloud maps with dense image pixels, which inevitably introduces depth errors and degrades pose estimation accuracy. We propose a monocular visual odometry framework utilizing a continuous 3D Gaussian map, which directly assigns geometrically consistent depth values to all extracted high-gradient points without interpolation. Evaluations on two public datasets demonstrate superior tracking accuracy compared to existing methods.
	We have released the source code of this work for the development of the community.
\end{abstract}

\section{Introduction}
\label{Introduction}

Visual odometry (VO)/visual-inertial odometry (VIO) is a crucial capability in a wide range of technologies, including robotics, unmanned aerial vehicles and mixed reality. However, the simultaneous estimation of camera poses and 3D map points is a highly non-convex optimization problem, limiting the general applicability of VO and VIO. The introduction of prior maps \cite{yuan2023sdv, yuan2024sr, yuan2024srlivo} allows VO and VIO to avoid the estimation of 3D map points, thereby significantly reducing the dimensionality of optimization variables. This structural simplification mitigates non-convexity in the objective function, consequently improving the accuracy and robustness of systems.

\begin{figure}
	\begin{center}
		\includegraphics[scale=0.5]{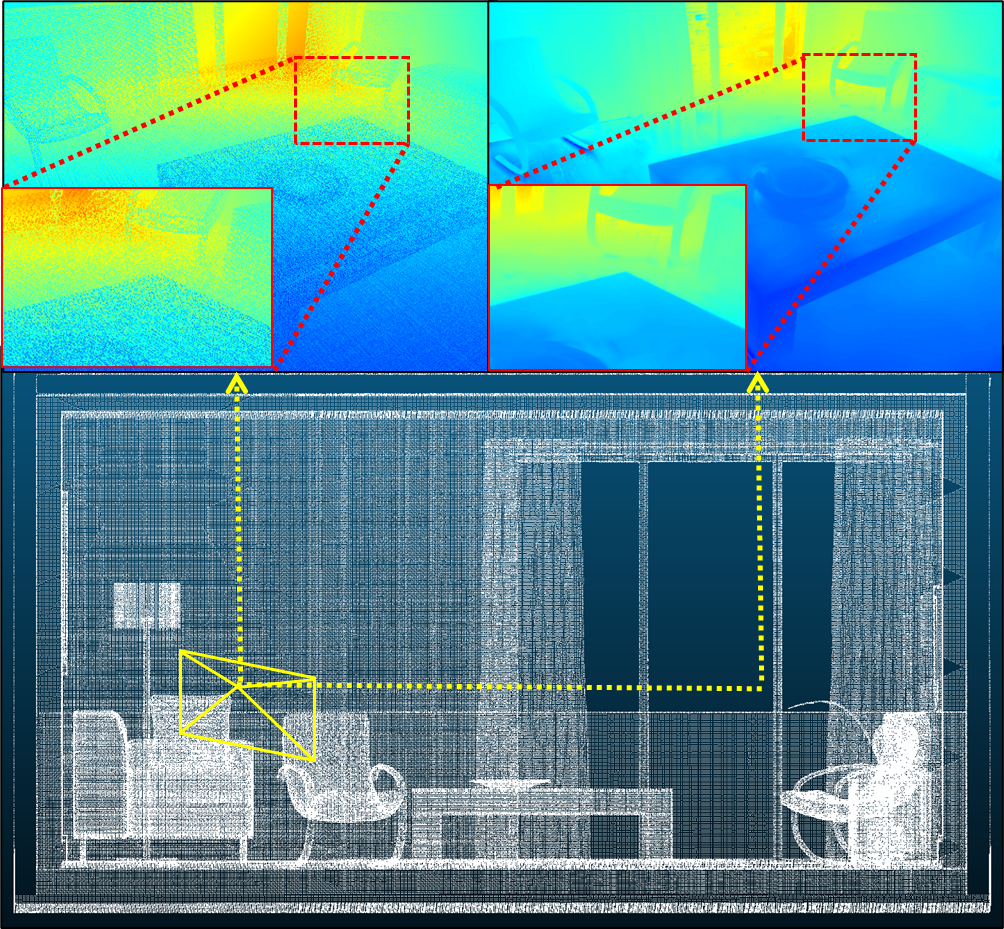}
		\caption{The top left picture and top right picture are depth maps from interpolation and gaussian splatting respectively. The bottom one is discrete point cloud map.While interpolation produces unreliable depth estimates in point cloud gaps, Gaussian splatting maintains consistent accuracy in these regions.}
		\label{fig1}
	\end{center}
\end{figure}

In map-based VO/VIO systems, the initial position of the local sensor in the prior map is determined either through relocalization or by predefined extrinsic parameters \cite{zuo2020multimodal}, \cite{wolcott2014visual}. Subsequently, depth values are assigned to image pixels by associating the prior map with local images which are assumed to be accurate enough for constructing optimization constraints. However, unlike dense images, the prior maps in most existing map-based VO/VIO approaches are discretely represented. Not all features or high-gradient regions detected in 2D images may reliably match corresponding 3D points in discrete prior maps.
To address this limitation, most existing approaches estimate the depth of 2D features or high-gradient points by approximating interpolation. Nevertheless, this approximation degrades the accuracy of the original data, as illustrated in Fig.\ref{fig1}. Furthermore, since these interpolated depth values propagate into optimization constraints, the interpolation errors accumulate across system modules, ultimately compromising overall performance.

In this paper, we propose a Gaussian map-based direct monocular visual odometry. The 3D Gaussian splatting technique can continuously represent the raw discrete prior map. By leveraging the continuous representation capability of 3D Gaussian Splatting \cite{kerbl20233d}, we convert the raw discrete prior map into a continuously dense map. This rendered map provides corresponding depth values for specific image positions. Since the pixels in the generated depth map have a one-to-one correspondence with pixels in the images, all features or high-gradient points can acquire their respective depth values without approximation. 
This approach avoids interpolation errors during data association, ensuring accurate and reliable depth assignments.

Experimental results on two public datasets demonstrate that continuous map has significant potential in enhancing the accuracy and reliability of VO/VIO. Moreover, 3D Gaussian rendering also enhances the visual interpretability of prior maps.

To conclude, the main contributions of this paper are mainly twofold:
1) We present a cross-modal visual odometry that utilizes a monocular camera and prior 3D Gaussian maps, providing continuous depth of the scene to estimate poses. The method achieves more accurate localization results than existing methods on two public datasets.
2) We have released the source code of our approach to facilitate the development of the community. \footnote{The source code is available at \url{https://github.com/JD-hust/gs-dso}}

The rest of this paper is structured as follows: Sec. \ref{Related Work} reviews the releted literature. Sec. \ref{Method} introduces our system, followed by experimental evaluation in Sec. \ref{Experiments}. Finally, Sec. \ref{Conclusion} summarizes the paper.

\section{Related Work}
\label{Related Work}

In this section, we provide an overview of previous researches on map-based VO/VIO and Gaussian splatting techonology for VSLAM. 

\subsection{map-based VO and VIO}

Prior map-based VO/VIO needs to establish associations from different sensors and construct constraints between 3D and 2D data. Based on the characteristics of the constraints established, these approaches can be primarily divided into two categories: spatial matching-based constraints and image plane-based constraints.

Methods with spatial matching-based constraints typically involve aligning point clouds reconstructed from visual odometry with LiDAR-derived 3D prior maps to estimate camera poses through spatial correspondence optimization. Forster et al.\cite{forster2013air} proposed a collaborative localization and mapping between ground-based depth sensors and aerial monocular cameras addressing the alignment problem of 3D surfaces from heterogeneous sensors. The pose is iteratively optimized through Iterative Closest Point (ICP) \cite{besl1992method} between 3D map points and reconstructed points. Following the idea of spatial matching, Caselitz et al. \cite{caselitz2016monocular} proposed a two-stage framework: 1) reconstructing sparse point clouds through local bundle adjustment using visual feature correspondences, followed by 2) performing ICP-based alignment between the reconstructed map and prior LiDAR point clouds to estimate scales and poses. Yabuuchi et al. \cite{yabuuchi2021visual} utilized the point-to-plane Iterative Closest Point \cite{chen1992object} to substitute ICP scheme. 
Zuo et al. \cite{zuo2019visual} optimized depth to refine visual point clouds and utilized Normal Distribution Transform (NDT) \cite{biber2003normal} registration, combining semi-dense stereo-generated maps with prior maps to establish constraints. These approaches approximately assume one-to-one correspondences between reconstructed points and prior map points; however, such correspondence may not hold in reality, and the resolution mismatch between 2D reconedstructed points and 3D map points can exacerbate this issue.

On the other hand, methods with image plane-based constraints transform or directly project 3D prior map information into 2D image space to establish data associations. Wolcott et al. \cite{wolcott2014visual} bypassed conventional engineering pipelines by synthesizing multi-perspective images from 3D prior maps and subsequently maximizing mutual information with real images for localization. Lu et al. \cite{lu2017monocular} manually extracted sparse landmark features (e.g., road markings) and performed chamfer matching against 3D point clouds, combining odometry and epipolar geometry constraints. Yu et al. \cite{yu2020monocular} proposed a method that matches offline extracted 3D lines from point clouds and online extracted 2D lines from images to facilitate cross-modal associations. Zhou et al. \cite{zhou2021visual} extended \cite{yu2020monocular} by integrating both point and line features for matching, while \cite{zheng2023tightly} advances the system of \cite{yu2020monocular} from a loosely-coupled to a tightly-coupled framework. 
These methods based on line features are more robust, however, due to the discrete nature of point clouds, their ability to represent linear structures is inferior to that of planar structures. Consequently, line features detected in 3D point clouds may not precisely correspond to 2D image line features. These inaccuracies, exacerbated by error propagation, significantly degrade pose estimation accuracy. Ye et al. \cite{ye2020monocular} processed prior maps by rendering vertex and normal maps and utilized coplanar constraints to associate prior maps with images. But there is also the issue of discrete depth initialization with image points. Kim et al. \cite{kim2018stereo} minimized depth residuals between depth maps generated by stereo cameras and prior map projection to estimate poses. Constructing residuals with depth also faces the error effect of interpolation at point cloud holes.

In the aforementioned methodologies, depth is employed to either construct system constraints or directly serve as an observation for estimating poses and scales. The discrete nature of 3D point cloud maps leads to incomplete association when performing cross-modal associations with dense 2D images. Interpolating missing depth values from sparse point cloud data has become a general approach to address this challenge in existing frameworks. However, the depth error introdueced still cause errors in subsequent computations of the whole system.

\subsection{Gaussian splatting for VSLAM}

3D Gaussian splatting \cite{kerbl20233d} is a differentiable map representation method that is capable of densely and continuously modeling environmental appearance and structure, demonstrating significant potential in SLAM. Some recent approaches primarily utilize Gaussians for online mapping and localization. 
SplaTAM \cite{keetha2024splatam} leverages RGB-D cameras to initialize Gaussian ellipsoids and estimates poses through photometric error backpropagation derived from rendering loss. MonoGS \cite{matsuki2024gaussian} utilizes ORB \cite{rublee2011orb} feature points to incorporate Gaussian ellipsoids and bases ORB-SLAM3 \cite{campos2021orb} for pose estimation. Gaussian-SLAM \cite{yugay2023gaussian} proposes differential depth rendering and frame-to-model alignment for localization. Nevertheless, in these online Gaussian mapping frameworks, the estimation of Gaussian maps and pose parameters mutually influence each other, while noise from raw data amplifies errors during computation.

\begin{figure*}[h]
	\begin{center}
		\includegraphics[scale=0.5]{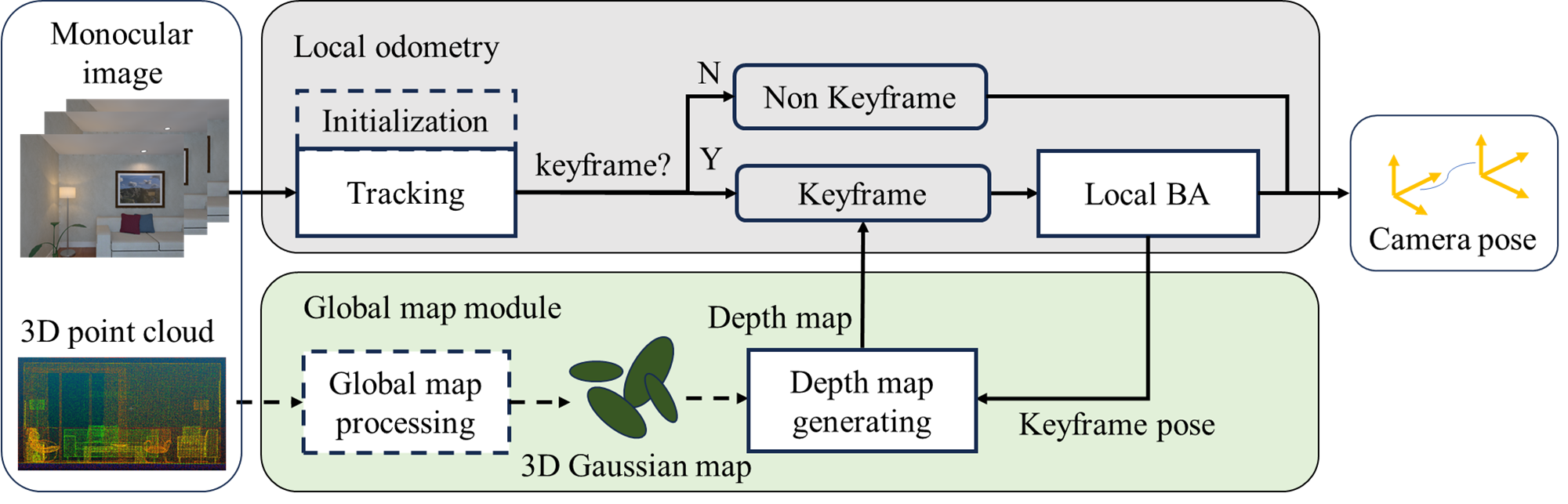}
		\caption{Overview of our system. The steps represented by dotted lines just run once: Initialization online and global map processing offline while the steps represented by solid lines execute as the system runs. }
		\label{fig2}
	\end{center}
\end{figure*}

In this paper, we aim to make the prior map representation continuous, thereby enabling more precise depth associations. We employ Gaussian splatting technology to render discrete map, generating Gaussian maps offline. Based on the benefits of direct visual tracking methods, we do not rely on explicit feature extraction. During the optimization process, we can update pixel correspondences between frames, and the depth maps from the Gaussian map can associate with RGB images without interpolation.

\section{Method}
\label{Method}

\subsection{System Overview}
\label{System Overview}

Fig. \ref{fig2} illustrates that our method framework consists of two modules: Global map module and Local odometry. We assume that the initial pose of the first frame in the global map is known. \textbf{Global map module}: Based on the poses obtained through local odometry, this module generates corresponding depth maps for each keyframe, enabling one-to-one depth associations with extracted high-gradient points. \textbf{Local odometry}: The initialization assigns an inverse depth value to the high-gradient points extracted from the first frame. After initialization, the system tracks all images sequentially. For frames identified as keyframes, their poses are passed to the global map module for depth information from prior maps.

\subsection{Global map module}
\label{Global map module}
Global map module contains two main parts: global map rendering and depth map generation. The former generates 3D Gaussian map from prior point clouds, which assigns the ability of continuous expression for the prior map, while the latter generates depth maps for each keyframe, which provides the depth information for the local odometry module.

For the first part, we train 3D Gaussian map following the explicit map construction method \cite{kerbl20233d}. The map consists of many 3D Gaussian ellipsoids and each Gaussian ellipsoid has the attributes: position, scale and covariance representing shape, color and opacity. All the attributes are optimized by loss iterations. The specific steps for this part are as follows:

1) 3D Gaussian initialization with LiDAR point cloud: The point cloud map is used for Gaussian initialization. For each point in the point cloud, a Gaussian ellipsoid is initially set with a random scale, covariance matrix and color on the point position, with an initial opacity value of 0.8. High opacity value prevents the background color from overly influencing the rendering effect. 

2) Loss computation and iteration: After initializing, images are generated by the following way:
\begin{equation}
\mathrm{\mathbf{c}_p=\sum_{i\in N}^n\mathbf{c}_i\alpha_i\prod_{j=1}^{i-1}(1-\alpha_j)}
\end{equation}
where $\mathrm{\mathbf{c}_p}$ represents the color of the generated image at pixel $\mathrm{p}$, $\mathrm{\mathbf{c}}_i$ is the color of the $\mathrm{i}$-th Gaussian ellipsoid, $\alpha_i$ is the opacity of the $\mathrm{i}$-th Gaussian ellipsoid, and $N$ is the set of the Gaussian ellipsoids that contribute to the pixel $\mathrm{p}$. The product term represents the accumulated transparency of the previous ellipsoids. 

The loss function is defined as:
\begin{equation}
\mathrm{l_c=\sum_{p\in P}||\mathbf{c}_p-\overline{\mathbf{c}}_p||^2}
\end{equation}
\begin{equation}
\mathrm{l=\lambda_{ssim}l_{ssim} + (1-\lambda_{ssim})l_c}
\end{equation}
where $\mathrm{P}$ is the set of all pixels in the generated image, $\overline{\mathbf{c}}_p$ is the color of the pixel $\mathrm{p}$ in the ground-truth image. $\mathrm{l_{ssim}}$ is the structure similarity index measure and $\mathrm{l_c}$ is total difference between true RGB values and generated values. The optimization process is conducted by minimizing the loss function. Through multiple iterations, these Gaussian ellipsoids are optimized to the shapes that fitting the geometric structure of scenario. Ideally, we can obtain a perfectly trained 3D Gaussian map. Sufficient training of the Gaussian map requires enough viewpoint observations, so we input the ground-truth images and camera poses for map training.

For the second part, global map module generates depth maps similar to generating colorful images:
\begin{equation}
\mathrm{d_p=\sum_{i\in N}^nz_i\alpha_i\prod_{j=1}^{i-1}(1-\alpha_j)}
\end{equation}
where $\mathrm{d_p}$ represents the depth value generated at pixel $\mathrm{p}$ and $\mathrm{z_i}$ represents the z-coordinate of the $\mathrm{i}$-th Gaussian ellipsoid in the camera coordinate system. When the system runs in real time, the part captures the keyframe poses from the local odometry, and projects the depth maps at the estimated poses.

This module enables the depth association without approximate interpolation. Point cloud maps exhibit significant spatial discontinuities, resulting in many regions where pixels lack corresponding 3D points during projection or back-projection. To address this, most methods interpolate depth by identifying the nearest points. However, high-gradient points or image features, which can lead to imprecise depth estimation due to the inherent ambiguity of edge point interpolation. In contrast, our method leverages a 3D Gaussian map to learn precise geometric structures, enabling continuous spatial representation and accurate depth inference for all pixels in the image.

\subsection{Local Odometry}
\label{Local Odometry}
\begin{figure}
	\begin{center}
		\includegraphics[scale=0.6]{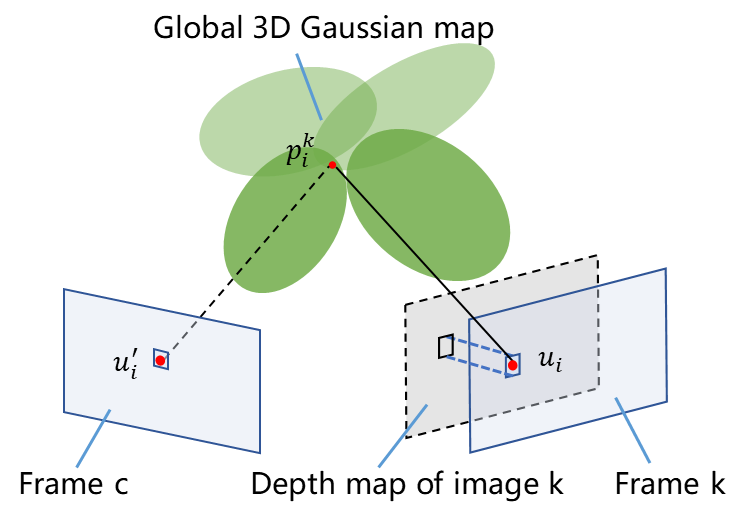}
		\caption{Illustration of depth association and projection relationship of our system. A tracking point is reconstrcuted from frame $k$ with the pixel-depth pair and then projected to frame $c$. The blue dotted line indicates the direct mapping relationship between depth maps and input images.}
		\label{fig3}
	\end{center}
\end{figure}
For each arrived new frame, the system extracts high-gradient points from the raw image data. Our system creates 3D points with high-gradient 2D pixels and their corresponding depth provided by the generated depth map from global map module. A 3D point created from frame $k$ is represented as:
\begin{equation}
	\mathrm{\mathbf{p}_i^{k}=\pi^{-1}(\mathbf{u}_i,\rho_i)}
\end{equation}
where $\mathrm{\mathbf{u}_i}$ is the 2D pixel on frame $k$ and $\mathrm{\rho_i}$ is the inverse depth value of $\mathrm{p_i^{k}}$. $\mathrm{\pi}$ represents the intrinsic matrix that transforms 3D points to 2D pixels.

The corresponding pixel of the point in another frame $c$ has the relationship:
\begin{equation}
	\mathrm{\mathbf{u}_i'=\pi(\mathbf{T}_{ck}\,\mathbf{p}_i^{k})}
\end{equation}
where $\mathrm{\mathbf{u}_i'}$ is the pixel position in frame $c$, and $\mathrm{\mathbf{T}_{ck}}$ is the transformation matrix that transforms points from frame $k$ to frame $c$.

For each new frame, our system first performs photometric transformation between the pixels of different frames, while the photometric error is computed by comparing the transformed photometry with the photometry of the current frame. By minimizing the photometric error of all tracking points, the relative pose of each new frame with respect to the latest keyframe is obtained:
\begin{equation}
\mathrm{\mathbf{T}_{ck}=\min_{\mathbf{T}_{ck}}\sum_{\mathbf{u}_{i}\in N_{c,k}}\mu_{u}\left\|I_{c}\left[\mathbf{u}_i'\right]-I_{k}[\mathbf{u}_{i}]\right\|}
\end{equation}
where $\mathrm{I_c}$ and $\mathrm{I_k}$ represents the light intensity of the pixel and $\mathrm{\mu_u}$ is the weight of each pixel to the photomatric error. After tracking successes, following the keyframe strategy in \cite{yuan2021rgb}, the system determines whether a new keyframe should be created.

If frames determined as keyframes, the absolute poses of them will be optimized in the sliding window. Each keyframe in the sliding window contains numberous pixel-depth pairs. A bundle adjustmented optimization is performed to refine a more accurate pose. The optimized pose is sent to the global map module for generating the corresponding depth map. The depth association and projection relationship is illustrated in Fig. \ref{fig3}. The generated depth map has the same size as the input frame, so each pixel in the new keyframe has a corresponding depth value. 

\section{Experiments}
\label{Experiments}

\subsection{Experimental Setup}
\label{Experimental Setup}
\begin{table}[]
	\begin{center}
		\caption{Information of the Test Sequences}
		\label{table1}
		\begin{tabular}{c|cp{1.5cm}<{\centering}p{0.6cm}<{\centering}p{0.6cm}<{\centering}c@{}} \hline
										& Name      & Scenario    & Frame   & Time(s)    & Length(m) \\ \hline
		\multirow{4}{1.3cm}{ICL-NUIM}   & $icl\_1$  & Living room & 1510   & 51         & 6.54      \\
										& $icl\_2$  & Living room & 967    & 33         & 2.05      \\
										& $icl\_3$  & Living room & 882    & 30         & 8.43      \\
										& $icl\_4$  & Living room & 1242   & 42         & 11.32     \\
		\hline
		\multirow{4}{1.3cm}{Augmented ICL-NUIM}  & $augm\_1$ & Living room & 2830   & 96         & 37.2      \\
												 & $augm\_2$ & Living room & 2350   & 78         & 36.8      \\
												 & $augm\_3$ & Office     & 2690   & 90         & 32.1      \\
												 & $augm\_4$ & Office     & 2538   & 85         & 38.1      \\
		\hline
		\end{tabular}
	\end{center}
\end{table}

\begin{table*}[]
	\begin{center}
	\caption{Localization Performance Comparsion}
	\label{table2}
	\begin{threeparttable}
	\begin{tabular}{c|cccc|cccc|cccc}
	\hline
								  & \multicolumn{4}{c|}{RMSE of ATE(m)}                               						  & \multicolumn{4}{c|}{RMSE of RTE(m)}                                                       & \multicolumn{4}{c}{RMSE of RRE(deg)}                                              		   \\  \cline{2-13} 
	\multirow{-2}{*}{Sequences}   & DSL                	 & RGBD-DSO             & MonoGS               & Ours                 & DSL                  & RGBD-DSO             & MonoGS 		 	   & Ours                 & DSL  				 & RGBD-DSO             & MonoGS 			   & Ours                  \\  \hline
	$icl\_1$                      & \uline{0.185}        & 0.346                & 0.044                & \textbf{0.027}       & 0.017                & \uline{0.006}        & 0.027                & \textbf{0.002}       & 0.405                & \uline{0.158}        & 1.258                & \textbf{0.108}                 \\
	$icl\_2$                      & 0.047                & \uline{0.010}        & 0.148                & \textbf{0.008}       & 0.006                & \textbf{0.001}       & 0.072                & \textbf{0.001}       & 0.130                & \textbf{0.020}       & 2.237                & \uline{0.026}                 \\
	$icl\_3$                      & \textbf{0.009}       & 0.064                & 0.278                & \uline{0.012}        & \textbf{0.001}       & 0.002                & 0.038                & \textbf{0.001}       & \uline{0.035}        & 0.043                & 1.044                & \textbf{0.022}        \\
	$icl\_4$                      & 0.281                & \uline{0.088}        & 0.996                & \textbf{0.029}       & 0.022                & \textbf{0.002}       & 0.064                & \uline{0.003}    	  & 0.668                & \uline{0.105}        & 1.615                & \textbf{0.183}        \\
	$augm\_1$                     & 1.534                & \uline{1.258}        & 1.396                & \textbf{1.239}       & 0.031                & \textbf{0.009}       & 0.134                & \uline{0.012}        & 0.924                & \textbf{0.216}       & 4.214                & \uline{0.259}         \\
	$augm\_2$                     & 0.020                & \textbf{0.014}       & 1.310                & \textbf{0.014}       & \textbf{0.002}       & \textbf{0.002}       & 0.162                & \textbf{0.002}       & 0.052                & \uline{0.038}        & 3.384                & \textbf{0.035}        \\
	$augm\_3$                     & $\times$             & \textbf{1.382}       & \uline{1.481}        & 1.519                & $\times$             & \uline{0.016}        & 0.178                & \textbf{0.008}       & $\times$             & \uline{0.830}        & 4.695                & \textbf{0.223}        \\
	$augm\_4$                     & \uline{0.055}        & 0.061                & 2.282                & \textbf{0.043}       & 0.012                & \uline{0.004}        & 0.141                & \textbf{0.002}       & 0.305                & \uline{0.146}        & 5.157                & \textbf{0.067}        \\ \hline
	\end{tabular}
	\begin{tablenotes}
        \footnotesize
        \item \textbf{Denotations}: "$\times$" means the corresponding value is not available as the system crashed on the sequence. The best result is marked in bold and the second-best result is marked in underline.
      \end{tablenotes}
  	\end{threeparttable}
	\end{center}
\end{table*}

\begin{table*}[]
	\begin{center}
		\caption{Ablation Experiment}
		\label{table3}
			\begin{tabular}{c|p{2.2cm}<{\centering}c|p{2.2cm}<{\centering}c|p{2.2cm}<{\centering}c} \hline
								   & \multicolumn{2}{p{4.4cm}<{\centering}|}{RMSE of ATE(m)} & \multicolumn{2}{p{4.4cm}<{\centering}|}{RMSE of RTE(m)} & \multicolumn{2}{p{4.4cm}<{\centering}}{RMSE of RRE(deg)} \\   \cline{2-7} 
			\multirow{-2}{3cm}{Sequences\centering}& Interpolation   & Ours\centering       & Interpolation   & Ours            & Interpolation   & Ours        \\	\hline
			$icl\_1$                   & 0.650           & \textbf{0.027}  & 0.007           & \textbf{0.002}  & 0.139           & \textbf{0.108}  \\
			$icl\_2$                   & 0.087           & \textbf{0.008}  & \textbf{0.001}  & \textbf{0.001}  & \textbf{0.025}  & 0.026           \\
			$icl\_3$                   & 0.121           & \textbf{0.012}  & 0.002           & \textbf{0.001}  & 0.031           & \textbf{0.022}  \\
			$icl\_4$                   & 1.305           & \textbf{0.029}  & 0.010           & \textbf{0.003}  & 0.503           & \textbf{0.183}  \\
			$augm\_1$                  & 2.088           & \textbf{1.239}  & \textbf{0.010}  & 0.012           & 0.318           & \textbf{0.259}  \\
			$augm\_2$                  & 0.529           & \textbf{0.014}  & 0.009           & \textbf{0.002}  & 0.264           & \textbf{0.035}  \\
			$augm\_3$                  & 2.337           & \textbf{1.519}  & 0.013           & \textbf{0.008}  & 0.277           & \textbf{0.223}  \\
			$augm\_4$                  & $\times$        & \textbf{0.043}  & $\times$        & \textbf{0.002}  & $\times$        & \textbf{0.067}  \\ \hline
			\end{tabular}
			\begin{tablenotes}
				\footnotesize
				\item \textbf{Denotations}: "$\times$" means the corresponding value is not available as the system crashed on the sequence. The best result is marked in bold.
			\end{tablenotes}
	\end{center}
\end{table*}

\textbf{Experimental Datasets.} Our evaluation protocol employs the ICL-NUIM \cite{handa2014benchmark} and Augmented ICL-NUIM \cite{choi2015robust} benchmarks, adhering to strict benchmark selection criteria that prioritize poses in global coordinate systems. From these datasets, eight sequences across two scenarioes were selected for systematic quantitative analysis. Four ICL-NUIM sequences were excluded from testing due to lack the groud-truth trajectory in global coordinate systems. The sequence characteristics are detailed in Table \ref{table1}.

\textbf{Evaluation metric and Computational Platform.} We implement three principal error metrics for systematic trajectory evaluation: Absolute Trajectory Error (ATE), Relative Translation Error (RTE), and Relative Rotation Error (RRE), all quantified through rigorous RMSE analysis. All the experiments are conducted on a consumer-level computer equipped with Intel Core i7-13700F processor with 32GB memory. The global map module is accelerated by NVIDIA GeForce GTX 1660Ti. 

\subsection{Localization performance}
\label{Localization performance}

In this section, we compare our system with the baseline RGBD-DSO \cite{yuan2021rgb}, another map-based system DSL \cite{ye2020monocular}, and an online 3D Gaussian map construction and localization system MonoGS \cite{matsuki2024gaussian}. For a fair comparison, we evaluate the performance of these methods by running the open-source codes provided by their offical implementations on our computational platform. For map-based systems (Ours/DSL), initialization leverages known-view pose priors to obtain initial depth association. To test on DSL, we followed the surfel map reconstruction description in DSL, generating density-calibrated surfel maps of the two scenarios. RGB-DSO evaluations utilize the RGB-D pairs provided by the datasets, where the depth images contain simulated noise. For MonoGS, trajectories undergo metric-scale recovery using least-squares alignment with ground-truth reference trajectories.

Quantitative benchmarking results in Table \ref{table2} reveal that our method demonstrates superior accuracy across 6/8 sequences on RMSE of ATE, RTE and RRE. On sequences where the best results were not achieved, our method either ranked second or maintained a marginal difference from the top performance. Catastrophic failure instances denoted by "$\times$" predominantly occurr in low-texture regions for DSL; It can also be demonstrated that our method is more robust than the other methods.

\subsection{Ablation Experiment}
\label{Ablation Experiment}

\begin{figure*}[]
	\centering
	\subfloat[]{
		\includegraphics[width=2.1cm, height=7.7cm]{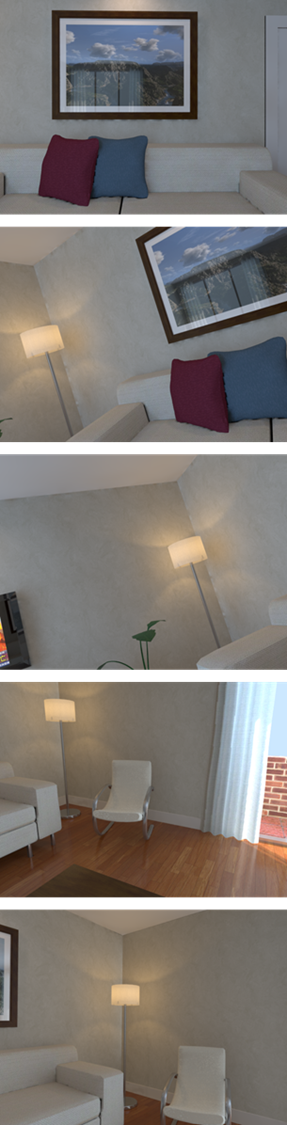}
		\label{5a}
	}\hspace{-3mm}
	\subfloat[]{
		\includegraphics[width=2.1cm, height=7.7cm]{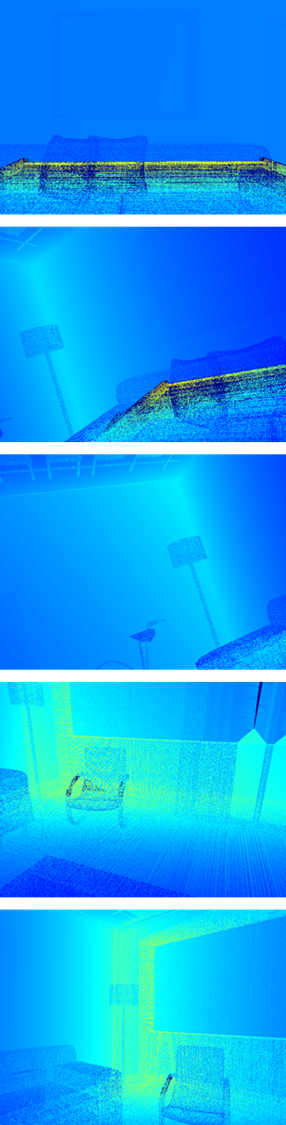}
		\label{5b}
	}\hspace{-3mm}
	\subfloat[]{
		\includegraphics[width=2.1cm, height=7.7cm]{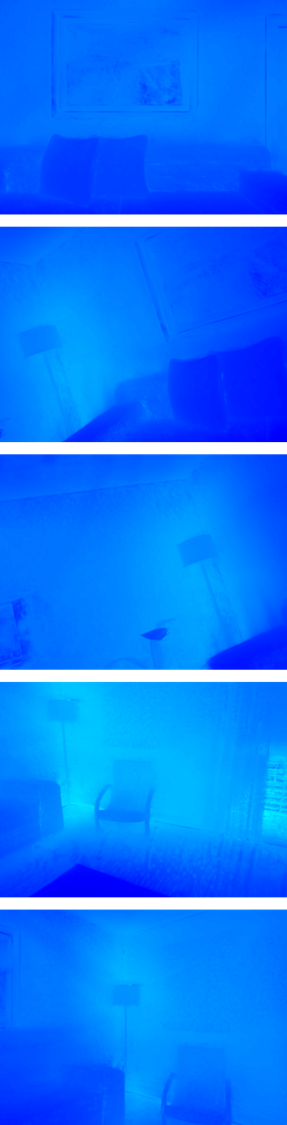}
		\label{5c}
	}\hspace{-3mm}
	\subfloat[\label{Ours}]{
		\includegraphics[width=2.1cm, height=7.7cm]{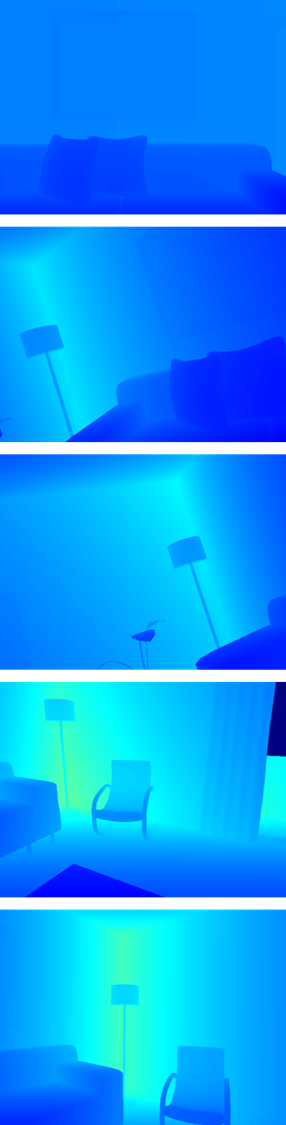}
		\label{5d}
	}\hspace{-2mm}
	\subfloat[]{
		\includegraphics[width=2.1cm, height=7.7cm]{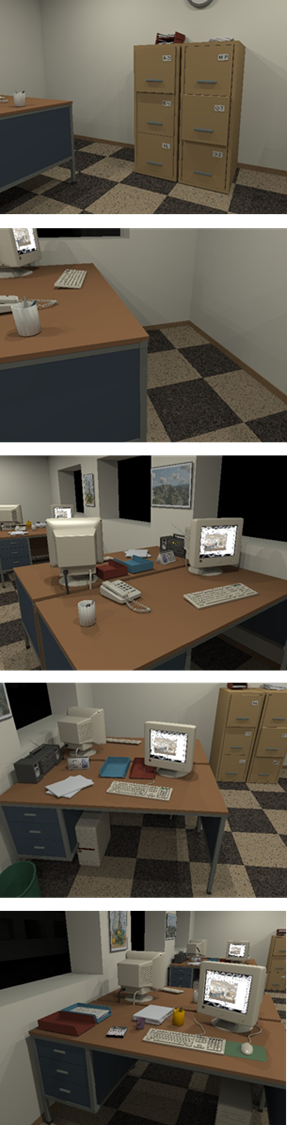}
		\label{5e}
	}\hspace{-3mm}
	\subfloat[]{
		\includegraphics[width=2.1cm, height=7.7cm]{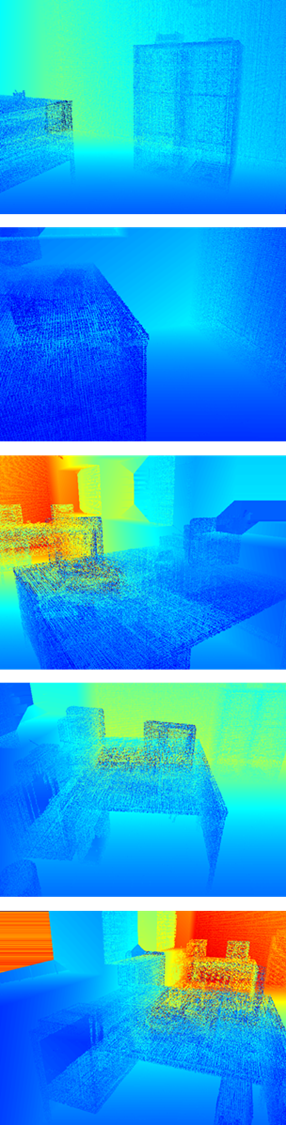}
		\label{5f}
	}\hspace{-3mm}
	\subfloat[]{
		\includegraphics[width=2.1cm, height=7.7cm]{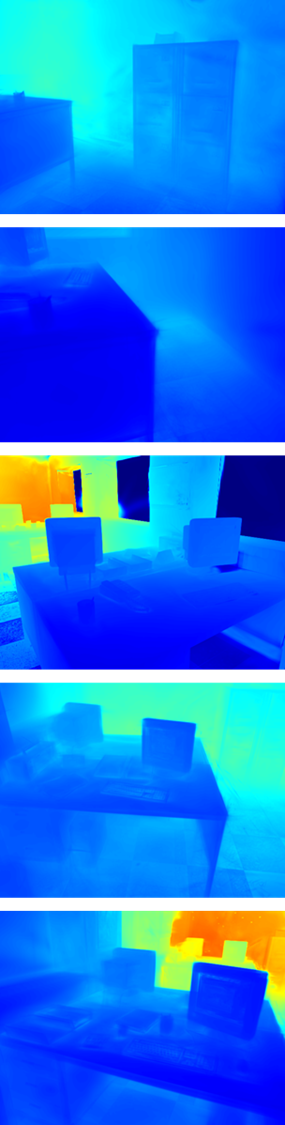}
		\label{5g}
	}\hspace{-3mm}
	\subfloat[]{
		\includegraphics[width=2.1cm, height=7.7cm]{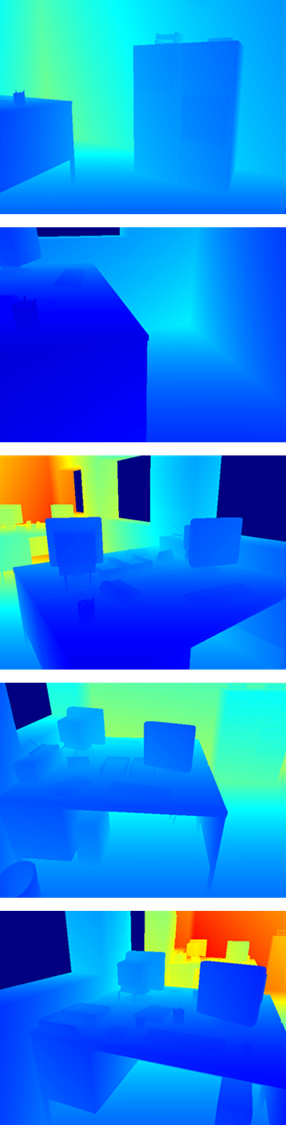}
		\label{5h}
	}
	\caption{(a)-(d),(e)-(h) examplar frames in seqence $icl\_1$ and $augm\_4$, respectively.(a)(e) are input images, (b)(f) are depth maps obtained by interpolation, (c)(g) are depth maps generated by our system and (d)(h) are ground-truth depth maps. They show the difference between the depth maps obtained using the interpolation method and the depth maps rendered from the Gaussian map. The dark areas in the picture set (b) represents depth values that are not valid due to interpolation mistakes.}
	\label{fig4}
\end{figure*}

Our architecture is predicated on the hypothesis that continuous volumetric representations inherently mitigate interpolation artifacts prevalent in discrete map-based localization. To validate this hypothesis, we conduct controlled ablation studies comparing two depth derivation paradigms: 1) Gaussian splatting-based continuous depth association and 2) conventional nearest-neighbor interpolation from discrete point cloud  representations. We integrated the depth interpolation module from the open-source code of \cite{zhang2014real}. The experimental configuration maintains identical frontend processing while substituting the depth mapping through modular replacement—activating either differentiable Gaussian renderer or the interpolation module.

As quantified in Table \ref{table3}, our continuous representation framework achieves statistically significant improvements across most all evaluation metrics. Crucially, the interpolation exhibits two critical failure modes: 1) Fundamental precision limitations, and 2) Depth-induced system instability causing catastrophic crashes. These results not only confirm the existence of interpolation-related error propagation mechanisms but also conclusively validate our theoretical framework's efficacy.

In addition, we visualize the depth calculated by the two different methods for the $icl\_1$ and $augm\_4$ sequence, comparing them with the ground-truth depth maps in Fig. \ref{fig4}. The interpolation approach exhibits geometrically inconsistent depth estimations in two critical areas: 1) Sparse regions containing point cloud voids, and 2) Occlusion boundaries with complex depth layering. These errors stem from the inherent limitations of discrete point cloud representations in modeling continuous surfaces and visibility constraints. In contrast, Gaussian splatting method achieves physically plausible depth reconstruction through opacity-aware ray accumulation, producing continuous depth maps that maintain structural consistency with ground-truth observations. The visual comparisons confirm the superior capability of our method in preserving geometric continuity and occlusion relationships compared to conventional interpolation techniques.

\subsection{Trajectory Visualization}
\label{Trajectory Visualization}

\begin{figure}
	\begin{centering}
	\subfloat[]{
		\includegraphics[width=3.8cm, height=3.25cm]{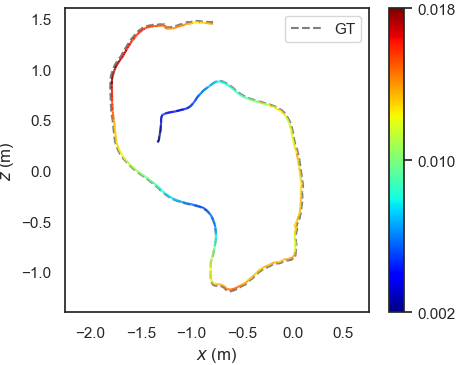} 
	}
	\subfloat[]{
		\includegraphics[width=3.8cm, height=3.25cm]{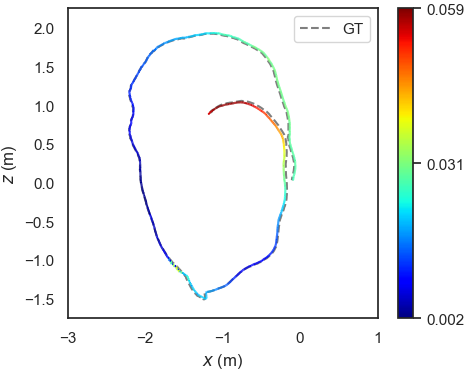} 
	}

	\subfloat[]{
		\includegraphics[width=3.75cm, height=3.2cm]{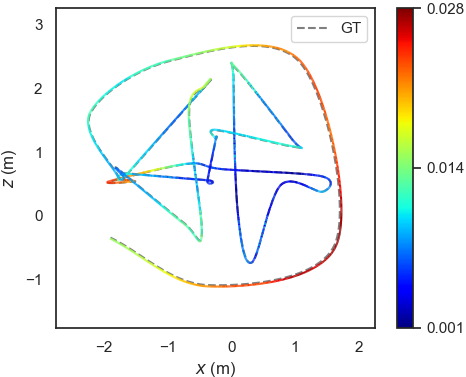} 
	}
	\subfloat[]{
		\includegraphics[width=3.75cm, height=3.2cm]{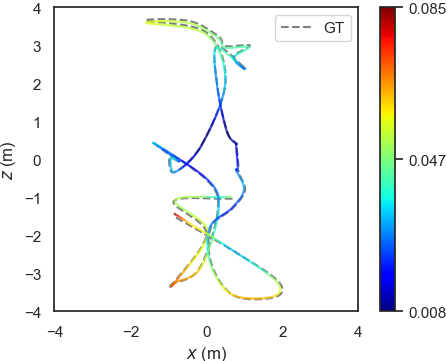}
	}
	\caption{(a) is the estimated trajectory of $icl\_2$, (b) is the estimated trajectory of $icl\_3$, (c) is the estimated trajectory of $augm\_2$ and (d) is the estimated trajectory of $augm\_4$.}
	\label{fig5}
	\end{centering}
\end{figure}

The estimated trajectories generated by our system are visualized across multiple test sequences. As demonstrated in Fig. \ref{fig5}, a comparative analysis between the estimated and ground-truth trajectories is presented for the $icl\_2$, $icl\_3$, $augm\_2$ and $augm\_4$ datasets. Visualization reveals minimal deviations between the estimated trajectories and their corresponding ground-truth references. As evidenced in the overlapping trajectory plots, estimation errors consistently remain within acceptable thresholds.

\section{Conclusion}
\label{Conclusion}

This study proposes a monocluar direct odometry framework for indoor environments that synergistitcally integrates prior 3D Gaussian representations. Our methodology employs differentiable 3D Gaussian primitives as geometric anchors within the map representation. The global map module generates physics-compliant depth maps through ray casting, enabling direct depth correspondence establishment between image observations and map geometry while eliminating conventional interpolation procedures. 

Built upon the RGBD-DSO \cite{yuan2021rgb} architecture, our technical contributions are primarily manifested in three aspects: 1) A hardware-agnostic localization paradigm that substitutes RGBD sensor dependency with photometric constraints derived from prior Gaussian maps; 2) Continuity-preserving depth estimation achieved through analytical Gaussian splatting, effectively mitigating interpolation-induced artifacts; 3) Superior positioning accuracy demonstrated through comprehensive benchmarks against conventional prior-map methods and the online reconstruction approach. Future investigations will focus on establishing a tightly bidirectional optimization framework between the odometry and map geometry refinement.

\normalem
\bibliographystyle{IEEEtrans}
\bibliography{IEEEabrv,IEEEExample}

\begin{thebibliography}{10}
\providecommand{\url}[1]{#1}
\csname url@rmstyle\endcsname
\providecommand{\newblock}{\relax}
\providecommand{\bibinfo}[2]{#2}
\providecommand\BIBentrySTDinterwordspacing{\spaceskip=0pt\relax}
\providecommand\BIBentryALTinterwordstretchfactor{4}
\providecommand\BIBentryALTinterwordspacing{\spaceskip=\fontdimen2\font plus
\BIBentryALTinterwordstretchfactor\fontdimen3\font minus \fontdimen4\font\relax}
\providecommand\BIBforeignlanguage[2]{{%
\expandafter\ifx\csname l@#1\endcsname\relax
\typeout{** WARNING: IEEEtran.bst: No hyphenation pattern has been}%
\typeout{** loaded for the language `#1'. Using the pattern for}%
\typeout{** the default language instead.}%
\else
\language=\csname l@#1\endcsname
\fi
#2}}

\bibitem{besl1992method}
P.~J. Besl and N.~D. McKay, ``Method for registration of 3-d shapes,'' in \emph{Sensor fusion IV: control paradigms and data structures}, vol. 1611.\hskip 1em plus 0.5em minus 0.4em\relax Spie, 1992, pp. 586--606.

\bibitem{biber2003normal}
P.~Biber and W.~Stra{\ss}er, ``The normal distributions transform: A new approach to laser scan matching,'' in \emph{Proceedings 2003 IEEE/RSJ International Conference on Intelligent Robots and Systems (IROS 2003)(Cat. No. 03CH37453)}, vol.~3.\hskip 1em plus 0.5em minus 0.4em\relax IEEE, 2003, pp. 2743--2748.

\bibitem{campos2021orb}
C.~Campos, R.~Elvira, J.~J.~G. Rodr{\'\i}guez, J.~M. Montiel, and J.~D. Tard{\'o}s, ``Orb-slam3: An accurate open-source library for visual, visual--inertial, and multimap slam,'' \emph{IEEE Transactions on Robotics}, vol.~37, no.~6, pp. 1874--1890, 2021.

\bibitem{caselitz2016monocular}
T.~Caselitz, B.~Steder, M.~Ruhnke, and W.~Burgard, ``Monocular camera localization in 3d lidar maps,'' in \emph{2016 IEEE/RSJ International Conference on Intelligent Robots and Systems (IROS)}.\hskip 1em plus 0.5em minus 0.4em\relax IEEE, 2016, pp. 1926--1931.

\bibitem{chen1992object}
Y.~Chen and G.~Medioni, ``Object modelling by registration of multiple range images,'' \emph{Image and vision computing}, vol.~10, no.~3, pp. 145--155, 1992.

\bibitem{choi2015robust}
S.~Choi, Q.-Y. Zhou, and V.~Koltun, ``Robust reconstruction of indoor scenes,'' in \emph{Proceedings of the IEEE conference on computer vision and pattern recognition}, 2015, pp. 5556--5565.

\bibitem{forster2013air}
C.~Forster, M.~Pizzoli, and D.~Scaramuzza, ``Air-ground localization and map augmentation using monocular dense reconstruction,'' in \emph{2013 IEEE/RSJ International Conference on Intelligent Robots and Systems}.\hskip 1em plus 0.5em minus 0.4em\relax IEEE, 2013, pp. 3971--3978.

\bibitem{handa2014benchmark}
A.~Handa, T.~Whelan, J.~McDonald, and A.~J. Davison, ``A benchmark for rgb-d visual odometry, 3d reconstruction and slam,'' in \emph{2014 IEEE international conference on Robotics and automation (ICRA)}.\hskip 1em plus 0.5em minus 0.4em\relax IEEE, 2014, pp. 1524--1531.

\bibitem{keetha2024splatam}
N.~Keetha, J.~Karhade, K.~M. Jatavallabhula, G.~Yang, S.~Scherer, D.~Ramanan, and J.~Luiten, ``Splatam: Splat track \& map 3d gaussians for dense rgb-d slam,'' in \emph{Proceedings of the IEEE/CVF Conference on Computer Vision and Pattern Recognition}, 2024, pp. 21\,357--21\,366.

\bibitem{kerbl20233d}
B.~Kerbl, G.~Kopanas, T.~Leimk{\"u}hler, and G.~Drettakis, ``3d gaussian splatting for real-time radiance field rendering.'' \emph{ACM Trans. Graph.}, vol.~42, no.~4, pp. 139--1, 2023.

\bibitem{kim2018stereo}
Y.~Kim, J.~Jeong, and A.~Kim, ``Stereo camera localization in 3d lidar maps,'' in \emph{2018 IEEE/RSJ International Conference on Intelligent Robots and Systems (IROS)}.\hskip 1em plus 0.5em minus 0.4em\relax IEEE, 2018, pp. 1--9.

\bibitem{lu2017monocular}
Y.~Lu, J.~Huang, Y.-T. Chen, and B.~Heisele, ``Monocular localization in urban environments using road markings,'' in \emph{2017 IEEE intelligent vehicles symposium (IV)}.\hskip 1em plus 0.5em minus 0.4em\relax IEEE, 2017, pp. 468--474.

\bibitem{matsuki2024gaussian}
H.~Matsuki, R.~Murai, P.~H. Kelly, and A.~J. Davison, ``Gaussian splatting slam,'' in \emph{Proceedings of the IEEE/CVF Conference on Computer Vision and Pattern Recognition}, 2024, pp. 18\,039--18\,048.

\bibitem{rublee2011orb}
E.~Rublee, V.~Rabaud, K.~Konolige, and G.~Bradski, ``Orb: An efficient alternative to sift or surf,'' in \emph{2011 International conference on computer vision}.\hskip 1em plus 0.5em minus 0.4em\relax Ieee, 2011, pp. 2564--2571.

\bibitem{wolcott2014visual}
R.~W. Wolcott and R.~M. Eustice, ``Visual localization within lidar maps for automated urban driving,'' in \emph{2014 IEEE/RSJ International Conference on Intelligent Robots and Systems}.\hskip 1em plus 0.5em minus 0.4em\relax IEEE, 2014, pp. 176--183.

\bibitem{yabuuchi2021visual}
K.~Yabuuchi, D.~R. Wong, T.~Ishita, Y.~Kitsukawa, and S.~Kato, ``Visual localization for autonomous driving using pre-built point cloud maps,'' in \emph{2021 IEEE Intelligent Vehicles Symposium (IV)}.\hskip 1em plus 0.5em minus 0.4em\relax IEEE, 2021, pp. 913--919.

\bibitem{ye2020monocular}
H.~Ye, H.~Huang, and M.~Liu, ``Monocular direct sparse localization in a prior 3d surfel map,'' in \emph{2020 IEEE International Conference on Robotics and Automation (ICRA)}.\hskip 1em plus 0.5em minus 0.4em\relax IEEE, 2020, pp. 8892--8898.

\bibitem{yu2020monocular}
H.~Yu, W.~Zhen, W.~Yang, J.~Zhang, and S.~Scherer, ``Monocular camera localization in prior lidar maps with 2d-3d line correspondences,'' in \emph{2020 IEEE/RSJ International Conference on Intelligent Robots and Systems (IROS)}.\hskip 1em plus 0.5em minus 0.4em\relax IEEE, 2020, pp. 4588--4594.

\bibitem{yuan2021rgb}
Z.~Yuan, K.~Cheng, J.~Tang, and X.~Yang, ``Rgb-d dso: Direct sparse odometry with rgb-d cameras for indoor scenes,'' \emph{IEEE Transactions on Multimedia}, vol.~24, pp. 4092--4101, 2021.

\bibitem{yuan2024srlivo}
Z.~Yuan, J.~Deng, R.~Ming, F.~Lang, and X.~Yang, ``Sr-livo: Lidar-inertial-visual odometry and mapping with sweep reconstruction,'' \emph{IEEE Robotics and Automation Letters}, 2024.

\bibitem{yuan2024sr}
Z.~Yuan, F.~Lang, T.~Xu, and X.~Yang, ``Sr-lio: Lidar-inertial odometry with sweep reconstruction,'' in \emph{2024 IEEE/RSJ International Conference on Intelligent Robots and Systems (IROS)}.\hskip 1em plus 0.5em minus 0.4em\relax IEEE, 2024, pp. 7862--7869.

\bibitem{yuan2023sdv}
Z.~Yuan, Q.~Wang, K.~Cheng, T.~Hao, and X.~Yang, ``Sdv-loam: Semi-direct visual--lidar odometry and mapping,'' \emph{IEEE Transactions on Pattern Analysis and Machine Intelligence}, vol.~45, no.~9, pp. 11\,203--11\,220, 2023.

\bibitem{yugay2023gaussian}
V.~Yugay, Y.~Li, T.~Gevers, and M.~R. Oswald, ``Gaussian-slam: Photo-realistic dense slam with gaussian splatting,'' \emph{arXiv preprint arXiv:2312.10070}, 2023.

\bibitem{zhang2014real}
J.~Zhang, M.~Kaess, and S.~Singh, ``Real-time depth enhanced monocular odometry,'' in \emph{2014 IEEE/RSJ International Conference on Intelligent Robots and Systems}.\hskip 1em plus 0.5em minus 0.4em\relax IEEE, 2014, pp. 4973--4980.

\bibitem{zheng2023tightly}
X.~Zheng, W.~Wen, and L.-T. Hsu, ``Tightly-coupled line feature-aided visual inertial localization within lightweight 3d prior map for intelligent vehicles,'' in \emph{2023 IEEE 26th International Conference on Intelligent Transportation Systems (ITSC)}.\hskip 1em plus 0.5em minus 0.4em\relax IEEE, 2023, pp. 6019--6026.

\bibitem{zhou2021visual}
G.~Zhou, H.~Yuan, S.~Zhu, Z.~Huang, Y.~Fan, X.~Zhong, R.~Du, and J.~Gu, ``Visual localization in a prior 3d lidar map combining points and lines,'' in \emph{2021 IEEE International Conference on Robotics and Biomimetics (ROBIO)}.\hskip 1em plus 0.5em minus 0.4em\relax IEEE, 2021, pp. 1198--1203.

\bibitem{zuo2019visual}
X.~Zuo, P.~Geneva, Y.~Yang, W.~Ye, Y.~Liu, and G.~Huang, ``Visual-inertial localization with prior lidar map constraints,'' \emph{IEEE Robotics and Automation Letters}, vol.~4, no.~4, pp. 3394--3401, 2019.

\bibitem{zuo2020multimodal}
X.~Zuo, W.~Ye, Y.~Yang, R.~Zheng, T.~Vidal-Calleja, G.~Huang, and Y.~Liu, ``Multimodal localization: Stereo over lidar map,'' \emph{Journal of Field Robotics}, vol.~37, no.~6, pp. 1003--1026, 2020.

\end{thebibliography}

\end{document}